\let\TeXyear\year
\documentclass{IEEEaccess}

\let\setyear\year
\let\year\TeXyear

\usepackage{tikz}

\setyear{2024}

\usepackage{amsmath, amssymb, amsfonts}
\usepackage{algorithm, algorithmic}
\usepackage{graphicx}
\usepackage{textcomp}
\usepackage{cite}
\usepackage{multirow}
\usepackage{caption}
\usepackage{subcaption}
\usepackage{bm}
\usepackage{pgfplots}
\usepgfplotslibrary{fillbetween}
\pgfplotsset{compat=newest}
\usepackage{pgfplotstable}
\usepackage{url}

\definecolor{accessblue}{RGB}{0,105,154}

\def\BibTeX{{\rm B\kern-.05em{\sc i\kern-.025em b}\kern-.08em
    T\kern-.1667em\lower.7ex\hbox{E}\kern-.125emX}}

\begin{document}
\history{Date of publication xxxx 00, 0000, date of current version xxxx 00, 0000.}
\doi{XX.XXXX/ACCESS.XXXX.DOI}

\title{A Balanced Approach of Rapid Genetic Exploration and Surrogate Exploitation for Hyperparameter Optimization}

\author{\uppercase{Chul Kim}\authorrefmark{1} \uppercase{AND Inwhee Joe\authorrefmark{2}}}

\address[1]{Hanyang University, Seongdong-gu, Seoul, South Korea. 04763 (e-mail: ki4420@hanyang.ac.kr, ORCID: 0009-0002-8177-0869)}

\address[2]{Hanyang University, Seongdong-gu, Seoul, South Korea. 04763 (e-mail: iwjoe@hanyang.ac.kr, ORCID: 0000-0002-8435-0395)}

\markboth
{Kim and Joe \headeretal: }
{Kim and Joe \headeretal: }

\corresp{Corresponding author: Inwhee Joe (e-mail: iwjoe@hanyang.ac.kr).}

\begin{abstract}

This paper introduces a novel approach to hyperparameter optimization (HPO), proposing a methodology that balances exploration and exploitation to enhance optimization performance. While evolutionary algorithms (EAs) have shown potential in HPO, they often struggle with effective exploitation. To address this limitation, we propose an improved hyperparameter optimization (HPO) framework that integrates a linear surrogate model into the genetic algorithm (GA). The GA’s flexible structure allows for seamless integration of multiple optimization strategies, and the surrogate model significantly boosts its exploitation capabilities. Specifically, we achieved an average performance improvement of \textbf{1.89\%} (max \textbf{6.55\%}, min \textbf{-3.45\%}) over the existing state-of-the-art HPO strategy.

\end{abstract}


\titlepgskip=-15pt

\maketitle

\noindent
\textcopyright{} 2024 IEEE. Personal use of this material is permitted. However, permission to reprint/republish this material for advertising or promotional purposes or for creating new collective works for resale or redistribution to servers or lists, or to reuse any copyrighted component of this work in other works must be obtained from the IEEE.

\vspace{1ex}

This is the author's version of the paper that has been accepted for publication in IEEE Access. The final version of record is available at: \url{https://doi.org/10.1109/ACCESS.2024.3508269}

\vspace{2ex}

\section{Introduction}
\label{sec:introduction}
\PARstart{O}{ptimization} has been studied extensively across various fields for decades, playing a significant role in advancing human development. Its origins trace back to applications in engineering, economics, and transportation \cite{Zaheer2023}. In the late 19th century, optimization gained prominence in production management, leading to marked improvements in productivity. The 20th century witnessed a rapid acceleration in the development of optimization algorithms, fueled by advancements in computer technology. For instance, linear programming and integer programming emerged in the 1940s, revolutionizing industrial applications, while the Lagrange multiplier method introduced in the 1950s expanded the scope of constrained optimization research \cite{Sun2019, Gambella2021, Chow1997}.

Optimization is also a cornerstone in artificial intelligence (AI). Performance optimization maximizes model accuracy and stability through efficient learning, while resource optimization enhances computational efficiency by utilizing resources such as CPUs, GPUs, and memory effectively. Additionally, hyperparameter optimization (HPO) is a critical area of AI research, aimed at finding the optimal combination of variables to maximize model performance \cite{Ciccone2023}.

Hyperparameter optimization (HPO) is a critical step in enhancing the performance and efficiency of machine learning models. Various HPO techniques have been developed to tackle the challenges of optimizing hyperparameters in large-scale and computationally expensive scenarios. Among these, Bayesian Optimization (BO) has gained significant attention due to its ability to balance exploration and exploitation using probabilistic models. Recent advancements include the  Bayesian Optimization and HyperBand(BOHB) algorithm, which combines Bayesian optimization with Hyperband to improve efficiency and robustness for large-scale hyperparameter optimization tasks \cite{Falkner2018}. Hyperband is a resource-efficient method that leverages a bandit-based strategy to accelerate the optimization process. By iteratively allocating resources to promising configurations, Hyperband achieves faster convergence and improved scalability compared to traditional methods \cite{Li2018}.
Another noteworthy approach is the Tree-structured Parzen Estimator (TPE), which modifies Bayesian optimization by leveraging a probabilistic model to approximate the objective function. TPE has been extensively reviewed in the context of various algorithms and applications, highlighting its strengths and limitations \cite{Zhong2022}.
Evolutionary algorithms, such as Genetic Algorithms (GA), are also widely used in HPO due to their adaptability and capability to explore complex search spaces. GAs simulate natural selection and evolution to optimize hyperparameters, making them particularly effective for solving challenging optimization problems \cite{Yao2020}.
Reinforcement learning (RL)--based optimization represents a newer trend in HPO. RL-based methods dynamically adjust hyperparameters by modeling the optimization process as a decision-making problem. This approach has shown promise in automated machine learning (AutoML) settings, addressing state-of-the-art challenges in the field \cite{Mendoza2019}.
Finally, These methodologies illustrate the diverse and innovative strategies that have emerged in HPO, each contributing unique strengths to address different optimization challenges.

Among optimization techniques, Bayesian optimization has been a leading approach, especially in HPO, due to its efficient exploration of optimization spaces \cite{Cuevas2021}. This technique employs probabilistic models to iteratively select promising candidates for evaluation, balancing exploration and exploitation through the acquisition function \cite{Yu2010}. However, Bayesian optimization has notable limitations. First, employing complex models like Gaussian processes can lead to significant computational overhead, particularly as the number of hyperparameters grows. Second, tuning its hyperparameters, such as $\xi$ or $\kappa$, can introduce additional layers of complexity. Third, it struggles with maintaining a precise balance between exploration and exploitation. Finally, it may perform poorly when the objective function exhibits irregular patterns in the search space.

To address these challenges, this paper proposes a hybrid HPO approach that combines the strengths of genetic algorithms (GAs) and linear surrogate models. This method seeks to balance exploration and exploitation effectively while mitigating the drawbacks of Bayesian optimization.

The remainder of this paper is structured as follows. Section II discusses related work, focusing on existing HPO methods utilizing GAs and surrogate models. Section III presents our hybrid strategy, detailing the integration of GAs and linear surrogate models to enhance HPO performance. Section IV evaluates the proposed method using the HPOBench benchmark, demonstrating its superiority in complex optimization scenarios. Finally, Section V concludes the paper and outlines future research directions.

\section{Related Works}

\subsection{Genetic Algorithm (GA)}

GA is a computational algorithm that borrows the principles of biological evolution and heredity to approach the solution of optimization problems. Based on a randomly generated initial population, it adopts a strategy of repeatedly applying operations such as selection, crossover, and probabilistic mutation to derive the optimal solution over generations. GA tends to provide robust performance for a variety of optimization tasks \cite{Katoch2021}, but are characterized by the fact that they can require a significant number of function evaluations \cite{Kudela2022}. These GA effectively perform exploitation tasks through the crossover operation and enhance the exploration process through the probabilistic application of the mutation operation. These two major operations are related to the core capability of GA and play a crucial role in improving efficiency and effectiveness.
Genetic Algorithms (GAs) have been extensively used in Hyperparameter Optimization (HPO) due to their robust exploration capabilities in high-dimensional spaces. Recent advancements have demonstrated the effectiveness of GAs in improving machine learning model performance across various applications.
One notable application of GAs in HPO is for text analysis tasks. Researchers developed a GA-based hyperparameter tuning model that achieved high classification accuracy on datasets such as IMDB (88.73$\%$) and Yelp (92.17$\%$), while significantly reducing computational costs \cite{Saboori2024}. Similarly, GAs have been utilized for optimizing hyperparameters in Convolutional Neural Networks (CNNs), specifically for agricultural tasks such as pest detection. These studies demonstrated improved classification performance by fine-tuning transfer learning parameters \cite{Rana2023}.
Furthermore, GAs have been employed to optimize hyperparameters in deep learning models for PM2.5 concentration prediction. This approach outperformed default configurations and random search methods, achieving a reduction in Mean Squared Error (MSE) by 13.38$\%$ and 55.30$\%$ on test performance, respectively \cite{Wang2023a}. These results highlight the growing role of GAs in addressing complex optimization challenges in HPO.

\subsection{Surrogate-Assisted Evolutionary Algorithm(SAEA)}

Surrogate-Assisted Evolutionary Algorithm(SAEA) is a variant of evolutionary algorithms (EAs) that leverages surrogate models to replace computationally expensive fitness functions, enabling efficient exploration of optimal solutions at a reduced cost. By using surrogate models to approximate the fitness landscape, SAEA significantly lowers the computational burden associated with direct evaluations, making it an effective approach for solving complex optimization problems where high evaluation costs are a concern \cite{Jin2011}. Utilizing this surrogate model allows for more trials and reduces the number of fitness evaluations that could directly impact the system. This study proposes a structure that applies a surrogate model to the evolutionary algorithm to find better optimal solutions, which share similarities with SAEA. However, there are differences in how each component is utilized. The proposed method focuses on integrating the surrogate model into the basic evolutionary process of GA to enhance performance, offering a different optimization approach from that of SAEA in the way each element is employed.

\subsection{Surrogate Model}

Surrogate models are employed to reduce the computational cost of data-intensive tasks such as optimization and analysis by capturing the correlation between input and output, thus reducing complexity \cite{Viana2021, Srinivasan2022}.

These models enable rapid exploration of various parameters without the need for extensive individual calculations, aiding in the search for optimal solutions.

While surrogate models improve computational efficiency, they do not fully replace the original models and require a balance between accuracy and computation time.

\subsection{Hill climbing}

In Hill Climbing, an initial solution is chosen, and the algorithm explores neighboring solutions to improve the value of an objective function. The process continues until no further improvement is found, often leading to a local optimum \cite{HillClimbing2024}. As it attempts to continuously climb from the current location along the steepest slope to reach the top of the mountain, it has been likened to 'mountain climbing' \cite{Chinnasamy2022}.

One of the main drawbacks of the algorithm is that it is prone to being stuck in a local optimum instead of a global optimum \cite{Li2023}. This happens because the search stops when it reaches a solution that is only better than the surrounding solutions. Various modified algorithms have been proposed to overcome this problem. Stochastic Hill Climbing stochastically selects other solutions in addition to better ones, which can help prevent getting stuck in local optima \cite{Escamilla-Serna2022}. Random Restart Hill Climbing starts from several different initial solutions, running through multiple local optima to select the best solution among them. It was applied to optimize the flexible job shop scheduling problem, demonstrating that RRHC could refine solutions effectively by starting from various initial conditions \cite{Escamilla-Serna2022}. Simulated Annealing is a probabilistic global search algorithm that explores the solution space by accepting worse solutions with a certain probability, which decreases over time, allowing it to escape local optima and increase the likelihood of finding a global optimum \cite{Reihani2023}.

\subsection{Bayesian Optimization}

Bayesian optimization is a probabilistic global optimization method that models an objective function using a Gaussian process (GP) and iteratively updates the model with new observations. The approach balances exploration and exploitation via an acquisition function that selects the next point by considering potential improvement and uncertainty in the model \cite{Ghassemi2020}. At its core, it is an approach that considers the balance between exploration and exploitation and has the advantage of performing well even when evaluating complex objective functions or expensive functions \cite{Lee2024}.

Bayesian optimization is well suited to a wide range of machine learning problems because of its ability to incorporate a priori knowledge, ease of parallel computation, and probabilistic inference. However, despite these advantages, it still has several limitations. Model complexity increases as the high-dimensional search space expands, and the time required for optimization can be long. Randomness in the initial sampling process, the risk of local optima, and differences between the assumptions of the probabilistic model and reality are all factors that can degrade optimization performance \cite{Wang2023, Mei2022}. In addition, performance limitations in situations where parameter tuning is required or data is sparse must also be considered.

\subsection{HPOBench}

HPOBench is a benchmark framework for the evaluation of HPO algorithms using open datasets introduced at NeuraIPS 2021 \cite{Eggensperger2021}. It serves as a tool for evaluating HPO algorithms and provides standardized performance metrics using a large number of open datasets, providing an environment for users to compare and evaluate the performance of HPO algorithms. 

The main purpose of HPOBench is to evaluate the relative performance of HPO algorithms and select the optimal algorithm.  HPOBench provides a useful tool to perform a fair evaluation of HPO algorithms under the same conditions as possible by providing various benchmark datasets, machine learning and deep learning models, comparable HPO algorithms, and evaluation metrics through experiments, such as classification, regression, and image processing. 

\section{OUR APPROACHES: RAPID GENETIC EXPLORATION WITH RANDOM DIRECTION HILL-CLIMBING LINEAR EXPLOITATION(RGHL)}

Bayesian optimization attempts to strike a balance between exploration and exploitation\cite{Jasrasaria2019}. Exploration means examining areas that have not yet been evaluated, while exploitation means finding the best possible value based on current information. If too much emphasis is placed on the exploration phase, Bayesian optimization can get stuck in a local optimum by repeatedly exploring only the region around the local optimum. These optimization strategies make it difficult to measure the contribution of exploration and exploitation in detecting the optimal solution, and it is difficult to verify their performance as the equilibrium behavior progresses. Therefore, when developing our technique, we designed a structure that allows us to measure the contribution of exploration and exploitation, respectively. We developed the HPO framework, which allows these two strategies to contribute equally to reaching the optimal solution.

\subsection{Our Architecture}

\begin{figure*}[!t]
\centering
\includegraphics[width=\textwidth]{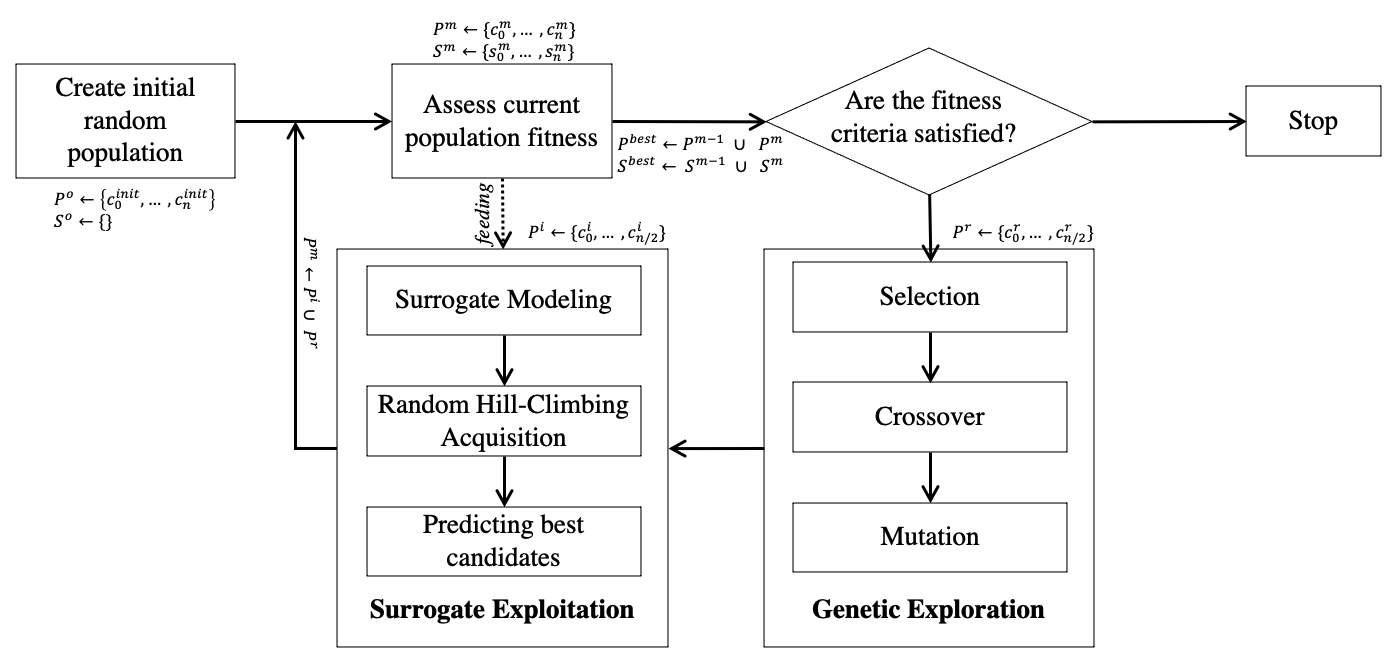}
\caption{Rapid Genetic Exploration with Random Direction Hill Climbing Linear Exploitation Architecture}
\label{fig2}                                                            
\end{figure*}

\begin{figure*}[!t]
\centering
$
\text{Chromosome} = \underset{type: mlp}{00000000|} \underset{batch: 16}{00000010|} \underset{lr: 1e-2}{00000010|} \underset{type: uniform}{00000011|} \underset{depth: 4}{00000100|} \underset{neurons: 8}{00000011|} \underset{opt: adam}{00000011|} \underset{act: relu}{00000001|} \underset{init: he}{00000000}
$
\caption{The figure represents how hyperparameters are stored using the chromosome structure that records genetic information in a GA. Each gene stores individual hyperparameter values.}
\end{figure*}

Figure 2 describes the structure of the chromosomes that make up a population in RGHL. Chromosomes are structures that store model types and their hyperparameters.

The RGHL architecture follows the basic structure of GA, a metaheuristic algorithm for solving optimization problems modeled on evolutionary theory.

We propose a structure that separates exploration and exploitation, inspired by Bayesian optimization's proportioning of exploration and exploitation in the process of finding new optimal solutions with an acquisition function so that each strategy contributes to achieving the objective in equal proportions without interfering with the other. The exploration strategy utilizes genetic algorithms to explore the unknown space through stronger genetic operations among observed individuals, while the exploitation strategy contributes to finding the optimal solution by utilizing surrogate model estimators to approximate the objective function.

Figure 1 represents the RGHL architecture in which the surrogate model estimator and genetic operator play the roles of exploitation and exploration, respectively, to find the optimal solution. In each generation, $50\%$ of the population is generated by rapid genetic operations focusing on the exploration strategy, and the remaining offspring are derived by the surrogate model dedicated to the exploitation strategy to form a new generation of the population. This design is to address the performance variations that arise depending on the value of $\xi$ of the expected improvement (EI) used in Bayesian optimization. In general, it is difficult to have insight into the ratio between exploration and exploitation during HPO as well as user-friendly guidelines.
By applying both strategies sequentially in equal proportions, the contribution of each strategy to the optimization process can be visualized, as shown in Figure 8, and the settings can remove parameters to reduce the burden on the user.

\subsection{Multi-Crossover and Multi-Mutation(Rapid Genetic Exploration)}

GA, which solves optimization problems modeled after evolutionary theory and Mendel’s laws of inheritance, are effective for general problems, but they have some drawbacks when used for HPO. The first is the convergence speed problem, which can consume a lot of time by stochastically changing a small number of genes in the process of repeating genetic operations.
This convergence speed problem is directly related to the expensive objective function call, so to compensate for this, we performed stronger genetic operations with multiple crossover and multiple mutations to generate many mutations per generation. This allows one to further explore the unknown search space.

%
%
\begin{figure}[h]
 \centering
  \includegraphics[width=\linewidth]{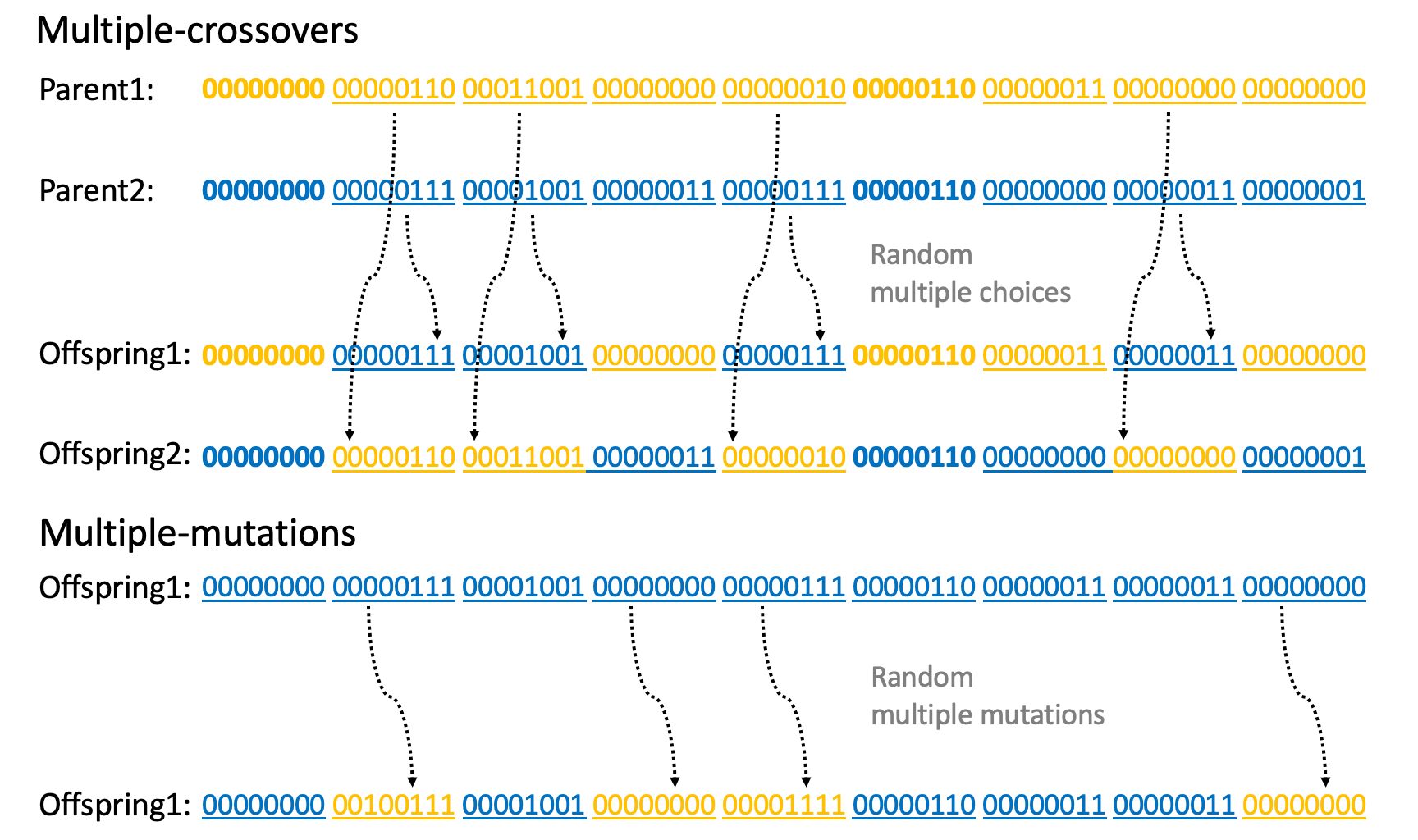}
 \caption{This figure represents Rapid Genetic Operations, which generate more crossover and mutation in a single generation to produce progressive offspring.}
\end{figure}

\begin{algorithm}[htbp]
\caption{Rapid Genetic Exploration}
\begin{algorithmic}[1]
    \renewcommand{\algorithmicrequire}{\textbf{parameter(s):}}
    \renewcommand{\algorithmicensure}{\textbf{output:}}
    \REQUIRE $F$ - Fitness function, $\alpha$ - Intensity factor of probability
    \ENSURE  Best individual
\\ \textit{Initialisation}
    \STATE $D$←Empty list 
    \COMMENT {Experience memory}
    \STATE $P$←RANDOM-POPULATION($P$)
    \STATE Store transition ($P$ , $F(P)$) in $D$
\\ \textit{Loop process} 
    \REPEAT
        \STATE $P'$←Empty list
        \FOR {i = 1 to SIZE(P)}
            \STATE $p_1$, $p_2$←TOP-N-RANDOM-CHOICES(P, 2)
            \COMMENT {$p$: Parent}
            \STATE $o_1$←$p_1$, $o_2$←$p_2$
            \COMMENT {$o$: Offspring}
            \STATE $n$←LENGTH($p_1$)
            \STATE $m$←Random number from 1 to $n$
            \STATE $c$←RANDOM-CHOICES([1:$n$], $m$)
            \FOR { $i$ = 1 to $n$}
                \IF {$i$ in $c$}
                    \STATE $o_1[i]$←$o_2[i]$, $o_2[i]$←$o_1[i]$
                \ENDIF
            \ENDFOR
            \COMMENT {Multi-crossover}
            \STATE $\beta$←SLOPE($D$)
            \COMMENT {Slope of fitness scores}
            \STATE $\delta$← $1.5 - {1}/{(1 + e^{-\alpha\cdot|\beta|})} $
            \COMMENT {Equation (1)}
            \STATE $r$←RANDOM()
            \IF {$\delta \ge r$} 
                \STATE $o_1$, $o_2$←MULTI-MUTATE($o_1$, $o_2$)
            \ENDIF
            \STATE add $o_1$, $o_2$ to $P'$
            \STATE Store transition ($P'$ , $F(P')$) in $D$
        \ENDFOR
        \STATE $P$←$P'$
    \UNTIL{Some individual is ﬁt enough, or enough time has elapsed}
    \RETURN Best individual in $P$
\end{algorithmic} 
\end{algorithm}

Algorithm 1 describes Rapid Genetic Exploration, which performs multiple crossovers and multiple mutations.

In this study, we change the genetic computation applied to HPO to emphasize exploration by applying a radical process to allow more crossover and mutation in a generation. To change a large number of genes in one generation, we changed the genetic algorithm to find the optimal value with an exploration strategy by performing multiple crossovers and multiple mutations. The selection step tried multiple crossovers and mutations by randomly selecting the top N parents sorted by fitness score.

\subsection{Adoptive Mutation Probability Function (AMPF)}

The problem with using GA for HPO is the local optima problem. Local optima can occur when the population consists of homogeneous chromosomes. In this case, the change in fitness values is minimal. The solution is to increase the probability of mutation to produce heterogeneous chromosomes to restore the population's fitness.

AMPF is a function that induces exploration by reducing the probability of mutation when the history of fitness scores obtained from the assessment of current population fitness during genetic operations grow monotonously and increasing the probability of mutation when there is no change in the score and dynamically changes the probability value according to the change in the slope value of the population fitness score history.

%
%
The adaptive Mutation Probability Function follows as:

\[(\hat{\beta_0}, \hat{\beta_1}) = \arg\min_{b_0, b_1} \sum^n_{i=1}(Y_i - b_0 - b_1 \cdot X_i)^2\]

\begin{equation}
f(Y, X) = 1.5 - \frac{1}{1 + e^{-\alpha \cdot |\hat{\beta_1}|}} \label{ampf} 
\end{equation}

Where $Y$ represents fitness scores of black-box function and $X$ represents sequences of scores(0:1) and $\hat{\beta_0}$ represents intercept of linear regression of scores and $\hat{\beta_1}$ represents the slope of linear regression of scores and $\alpha$ represents intensity factor of probability, when the value increases the likelihood increases stingy($1 \le \alpha, default=2$)

The AMPF value calculated by equation (1) is used as the probability of generating a mutation. It lowers the probability of mutation if the improvement in the population is monotonic, and increases the probability of more mutations if there is no trend.

%
%

\begin{figure}[!t]
\centerline{\includegraphics[width=\columnwidth]{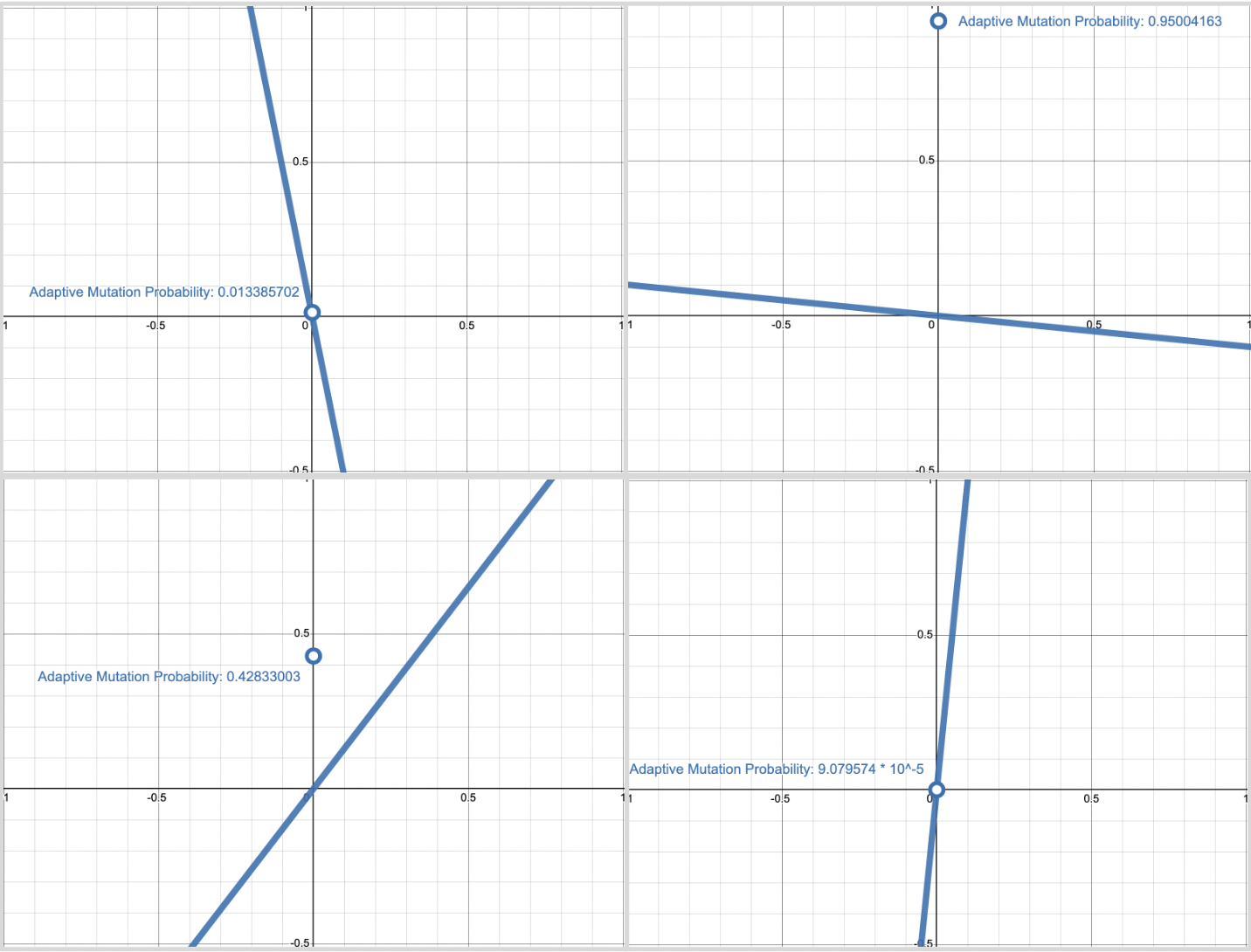}}
\caption{The Adaptive Mutation Probability Function (AMPF) represents the change in mutation probability as a function of slope in the fitness score history. It lowers the probability when the slope has a steep monotone, and increases the probability as the slope becomes smoother so that more mutations occur.( https://www.desmos.com/calculator/lhkqrgzqwn)}
\label{fig3}
\end{figure}

%
%
\begin{table}[h!]
\caption{95 Confidence Interval Evaluation of Adaptive Mutation Probability Function(AMPF) vs. Epsilon Greedy0.5(EG0.5) }
\label{xgb-perf-table}
\setlength{\tabcolsep}{3pt}
\begin{tabular}{|c|r|r|r|}
\hline
Algorithm& 
Test Loss& 
Valid Loss& 
Time(sec)\\
\hline

GA+AMPF&	0.17894±0.049293&	{\textbf{0.03291±0.013098}}&	133.35120±38.760646\\
GA+EG0.5&	0.16965±0.045888&	0.03923±0.016853&	120.33270±33.550151\\
\hline
\end{tabular}
\end{table}

We evaluated AMPF and ES0.5 as the mutation activity function of the GA in the HPOBench environment. Table 1 shows that GA+AMPF was evaluated to have better effective loss results and higher precision.

\subsection{Surrogate model}

We chose the surrogate model as a technique specialized for the exploitation strategy along with the exploration strategy of the GA described earlier. We propose a model that can estimate the optimal score by learning a surrogate model from the fitness scores ($Y$) and hyperparameters ($X$) obtained in the evaluation. After performing the genetic operation, the surrogate model learns the fitness scores obtained so far predicts the hyperparameter with the optimal score, and adds it as a population candidate. 
The ideal surrogate model should be designed as a single input multiple output (SIMO) structure, modeling a reversed black-box function that inputs the expected fitness score and outputs the optimal hyperparameter. With such a structure, exploitation can be effectively executed by utilizing only a surrogate model without an acquisition function. However, it is technically difficult to implement a reverse black-box function as a surrogate model because most HPOs are non-linear and stochastic, and the fitness score mapped to the hyperparameter in the search space is not unique.

%
%
\[X=\{x \in X^{M \times H}\}, Y=\{y \in Y^{N}\}, M \ge N\]
\[f(X)=Y\]
\begin{equation}
f^{-1}(Y)\not \neq X
\end{equation}

Equation (2) proves that the SIMO model cannot be implemented with the $f(X)$ function representing the black-box function, $X$ representing the hyperparameters, and $Y$ representing the black-box function score because the value of Score is not unique. To solve this problem, we can add identifiers to $X$ and $Y$ to implement a multiple input multiple output (MIMO) model.

%
%
\[T=\{ \tau \in T^M | \tau \sim U_{(0,1)}\}\]
\[X' = \{X, T\}, Y'=\{Y, T\}\]
\[X'=\{x \in X'^{M \times (H+1)}\}, Y'=\{y \in Y'^{M \times 2}\}\] 
\begin{equation}
f^{-1}(Y') = X'
\end{equation}

Equation (3) confirms that the MIMO model can be utilized to implement the reversed black-box function. We emphasize that the surrogate model of the reversed black-box function can be successfully constructed using $X^{\prime}$ and $Y^{\prime}$ with an identifier ($T$).
The implementation of the reversed black-box function indicates that the optimal hyperparameters can be effectively extracted without using the acquisition function. To build a MIMO model, various machine learning and deep learning techniques were utilized for training, and the suitability of the surrogate model was evaluated. In this evaluation process, the number of hyperparameters, loss value, and accuracy of the model are the main indicators to find the model that best fits the reversed black-box function.  Through these comparisons and evaluations, we ultimately found that the MLP model could most effectively fulfill the role of a surrogate model and selected it. We also conducted a comparative evaluation of multiple input single output(MISO), SIMO, and MIMO models to see how well they can achieve exploitation strategies in HPO environments.

\begin{figure}[h]
 \centering
 \includegraphics[width=\linewidth]{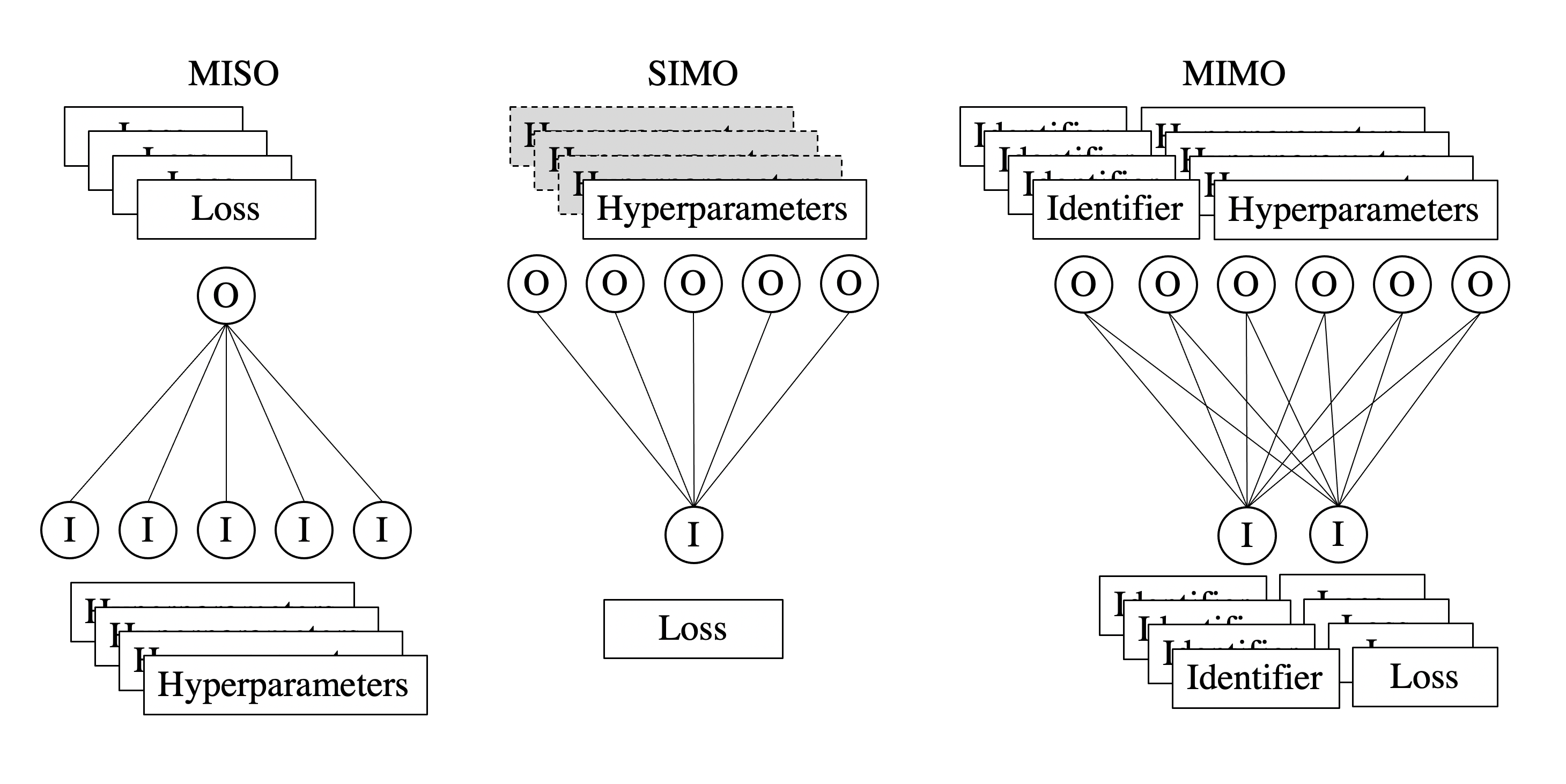}
 \caption{This figure explains the structure of MISO, SIMO, and MIMO. SIMO and MIMO are designed as reversed structures in which the inputs and outputs of the Score and Hyperparameter are reversed. SIMO can estimate one hyperparameter by inputting the expected score. The MIMO model can estimate multiple optimal hyperparameters separately by adding noise, and this noise is used as an identifier in the model.
}
\end{figure}

\begin{table}[h]
\caption{95 Confidence Interval Evaluations of MISO vs. SIMO vs. MIMO in HPOBench(XGBoost)}
\label{xgb-perf-table}
\setlength{\tabcolsep}{3pt}
\begin{tabular}{|c|r|r|r|}
\hline
Algorithm& 
Test Loss& 
Valid Loss& 
Time(sec)\\
\hline

MISO&	0.177359±0.00411&	{\textbf{0.036647±0.0069}}&	 428.088558±56.0205\\
SIMO&	0.201237±0.00753&	0.048144±0.0195&	 318.115392±21.3258\\
MIMO&	0.177677±0.00307&	0.047253±0.0077&	511.651570±62.6846\\
\hline
\end{tabular}
\end{table}

\subsection{Hill-climbing based Acquisition Function}

Our purpose is to develop a novel HPO outside of the Bayesian family, so a new acquisition function for the surrogate model is implemented.
Hill-climbing is a type of local search algorithm used to find the optimal solution to a particular problem. We implemented the acquisition function with it and developed Random direction Hill-Climbing(RHC), which is similar to Random Restart Hill-climbing.

%
%
\begin{algorithm}[htbp]
    \caption{Random direction Hill-Climbing(RHC)}
    \begin{algorithmic}[1]
        \renewcommand{\algorithmicrequire}{\textbf{parameter(s):}}
        \renewcommand{\algorithmicensure}{\textbf{output:}}
        \REQUIRE $Q$ - Surrogate model, $C$ - Number of genes, $V$ - Number of steps, $U$ - Best individuals
        \ENSURE  $B$ - Estimated best individuals
    \\ \textit{Initialisation}
        \STATE $B \gets$ Empty list, $R \gets$ Empty list
        \COMMENT{$R$: Fitness scores}
    \\ \textit{Loop process}
        \FOR {$u$ in $U$}
            \FOR {$j$ = 1 to $V$}
                \STATE $r^{*} \gets 0$
                \COMMENT{Initialize optimal score}  
                \STATE $d \gets$ RANDOM-INT-ARRAY(-1, 1, $C$)
                \COMMENT{Generate random directions}
    
                \REPEAT
                    \STATE $h \gets u - d$
                    \COMMENT{Climbing}
                    \STATE $r \gets Q(h)$
                    \IF {$r^{*} >=  r$} 
                        \STATE add $h$ to $B$, add $r$ to $R$
                        \STATE $u \gets h$, $r^{*} \gets r$
                    \ENDIF
                \UNTIL{$h == u$}
            \ENDFOR
        \ENDFOR
        \STATE $B \gets$ SORT-BY-ASC-SCORE($B$, $R$)
        \STATE $B \gets B[1:SIZE(U)] $
        \RETURN $B$
       
    \end{algorithmic} 
\end{algorithm}

Algorithm 2 presents a modified hill-climbing algorithm designed for use as the acquisition function in a surrogate model. The algorithm begins by generating random directions ($d$) by selecting gene values in the best individuals ($U$) from the range [-1,0,1]. It then subtracts these directions ($d$) from the best individual ($u$) to create new offspring. If the score ($r$) predicted by the surrogate model ($S$) shows improvement, the process continues. Otherwise, the algorithm progresses to the next iteration of the loop, generating a new random direction ($d$) and repeating the process. Once all iterations are completed, the algorithm returns the estimated number of top candidates, $B$.

%
%
\begin{algorithm}[htbp]
\caption{Random direction Hill-Climbing Linear Model(RHCLM)}
\begin{algorithmic}[1]
    \renewcommand{\algorithmicrequire}{\textbf{parameter(s):}}
    \renewcommand{\algorithmicensure}{\textbf{output:}}
    \REQUIRE $P$ - Population, $R$ - Fitness scores, $C$ - Number of Estimated individuals, $V$ - Number of steps
    \ENSURE  $L$ - Best estimated individuals
\\ \textit{Initialisation}
    \STATE $Q$←LINEAR-MODEL-Function with random weights $\theta$
    \STATE train $Q(P, R, \theta)$
    \COMMENT{Train surrogate model}
    \STATE $B \gets$ SORT-BY-ASC-SCORE($P$, $R$)
    \STATE $B \gets B[1:C]$
    \STATE $L  \gets RHC(Q_{\theta}, SIZE(B[1]), V, B)$
    \COMMENT{Algorithm 2}
    \RETURN $L$
\end{algorithmic} 
\end{algorithm}

Algorithm 3 outlines the implementation of the exploitation strategy, which involves training a Linear Surrogate Model and applying the Acquisition Function, as introduced in Algorithm 2. After testing in HPOBench's environments for boosting, neural networks, and SVMs, the linear model emerged as the most effective surrogate model. Notably, the linear model outperformed more advanced alternatives, indicating a potential risk of overfitting when using complex models in intricate exploration spaces. This observation aligns with Occam's razor, which favors simpler solutions when they perform comparably to more complex ones. Consequently, a simple yet robust low-dimensional linear model was selected as the final surrogate model.

%
%
\begin{algorithm}[htbp]
\caption{Rapid Genetic algorithm with random direction Hill-climbing Linear model(RGHL)}
\begin{algorithmic}[1]
    \renewcommand{\algorithmicrequire}{\textbf{parameter(s):}}
    \renewcommand{\algorithmicensure}{\textbf{output:}}
    \REQUIRE $F$ - Fitness function, $\psi$ - Number of Population, $\omega$ - Number of generations, $\xi$ - Number of best estimated individuals by RHCLM, $C$ - Num-
ber of Estimated individuals, $V$ - Number of steps
    \ENSURE  Best individual
\\ \textit{Initialisation}
    \STATE $D$← Empty list
    \COMMENT {Experience memory}
    \STATE $P$←RANDOM-POPULATION($\psi$ )
    \STATE Store transition ($P$, $F(P)$) in $D$
\\ \textit{Loop process} 
    \FOR { $i$ = 1 to $\omega$ }
        \STATE $P'$←Empty list
        \WHILE {SIZE($P'$) $\leq$ ($\psi$ - $\xi$)}
            \STATE $p_0$, $p_1$←TOP-N-RANDOM-CHOICES($P$)
            \STATE $o_0$, $o_1$←MULTI-CROSSOVER($p_0$, $p_1$)
            \STATE $o_0$, $o_1$←MULTI-MUTATE($o_0$, $o_1$)
            \COMMENT {With Equation (1)}
            \STATE add  $o_0$, $o_1$ to $P'$
        \ENDWHILE
        \COMMENT {Algorithm 1}
        \STATE $P$←$P'$, $R$←$F(P')$
        \STATE Store transition ($P$, $R$) in $D$
        \STATE $\pi$←RHCLM($D$, $C$, $V$)
        \COMMENT {Algorithm 3}
        \STATE add  $\pi$ to $P$
    \ENDFOR
    \RETURN Best individual in $P$
\end{algorithmic} 
\end{algorithm}

Algorithm 4 describes our proposed HPO algorithm, the Rapid GA with random direction Hill-climbing Linear model (RGHL), which is a synthesis of the Rapid GA (Algorithm 1), which is in the role of exploration, and Random direction Hill-Climbing Linear Surrogate Model (Algorithm 3), which is in the role of exploitation. The overall architecture and flowchart follow the GA and act equally as the exploration of genetic operations and the exploitation of the surrogate model to construct a generational population.

Initially, a uniformly distributed random population ($P$) is generated with $\psi$ number of generations and enters the generation ($P'$) loop. In a generation, first, perform a rapid genetic operation to generate $\frac{\psi}{2}$ number of populations, and then generate $\frac{\psi}{2}$ number of new offspring in the RHCLM step to perform population assessment ($F(P)$) with a black-box function. This Generation iterates over $\omega$ and returns the best value found by the process.

\section{Experiments}

We used HPOBench to evaluate the performance of RGHL. In this study, the performance of HPO algorithms is evaluated using specific benchmark tools within the benchmarks.ml environment: nn benchmark, rf benchmark, svm benchmark, and the xgboost package.

For each of the HPO algorithms selected for evaluation, the experiment is repeated with 300 black-box function calls a total of 100 times. Based on the results obtained, the performance is compared in terms of Test loss, Valid loss, and time taken for evaluation (in seconds). In particular, the valid loss value obtained from training the algorithm is utilized to calculate the fitness score of the black-box function.
The main evaluations in this study include the Rapid GA with hill-climbing Linear model (RGHL), which is newly proposed in this paper, and the Rapid GA with Hill-climbing Boosting model (RGHB), which utilizes a nonlinear surrogate model. In addition, the evaluation of existing popular HPO techniques such as Random Search, Bayesian Optimization, BOHB, and SMAC is also conducted at the same time. By doing so, it is aimed to verify the effectiveness and superiority of the proposed algorithm objectively.

Among the many datasets included within the benchmarks.ml framework, some large datasets are ruled out from the scope of our experiments. This is because processing large datasets and datasets containing many trait factors requires significant computational resources and time. Given this background, datasets from the OpenML-CC18 Curated Classification benchmark are selected for our experimental environment. From this benchmark, which contains a total of 72 datasets, 9 datasets are selected, excluding the large ones, to form the main test environment for our experiments. By leveraging this selection of datasets, we can maximize the efficiency of our experiments while still comprehensively covering the key aspects needed to evaluate the performance of our algorithms.

%
%
\begin{table}
\caption{OpenML-CC18 Classification benchmark dataset}
\label{openml-dataset-spec-table}
\setlength{\tabcolsep}{3pt}
\begin{tabular}{|c|l|r|r|r|}
\hline
Task-id& 
Name& 
\# of classes&
\# of features&
\# of instances\\
\hline
167151&	balance-scale&	3&	5&    625 \\
167154&	breast-w&	2&	10&    699 \\
167156&	mfeat-morphological&	10&	7& 2,000   \\
167163&	diabetes&	2&	9& 768 \\
167168&	vehicle&	4&	19& 846 \\
167187&	ilpd&	2&	11&    583 \\
167183&	banknote-authentication&	2&	5&  1,372   \\
167184&	blood-transfusion&	2&	5&   748 \\
167106&	climate-model&	2&	21&  540 \\
\hline
\end{tabular}
\end{table}

Table 3 presents the detailed characteristics of the datasets utilized in this experiment. Among the 72 datasets initially considered, datasets with more than 25 characteristic variables ($\#$ of features) or more than 2,000 instances of data ($\#$ of instances) were excluded from this experiment due to computational complexity and time efficiency considerations.
The main objective of this study is to evaluate the performance and validate the proposed RGHL algorithm. We tested RGHL, RGHB(a nonlinear surrogate model), and the following representative HPO techniques: Random Search, Bayesian Optimization, BOHB, and SMAC under the same conditions. Each HPO methodology experiments with exactly 300 calls to a black-box function on an open dataset, which allows for a detailed comparison and analysis of the optimum found by each HPO technique. In this way, it is aimed to evaluate the performance of RGHL in an objective and in-depth manner.

%
%
\begin{figure*}
    \centering
    \subfloat[XGBoost Benchmark]{
        \begin{tikzpicture}
            \begin{axis}[
                legend pos=north east,
                ymajorgrids=true,
                grid style=dashed,
                legend columns=2,
                legend style={font=\tiny},
                tick label style={font=\tiny},
                xlabel style={font=\tiny},
                width=0.28\textwidth,
                height=0.26\textwidth,
            ]
                \addplot[
                    color=blue,
                    opacity=1,
                    thick, 
                    smooth
                    ]
                    table [x={n}, y={sghl_min}] {xgb_results.dat};
                \addplot[
                    color=green,
                    opacity=1,
                    thick, 
                    smooth
                    ]
                    table [x={n}, y={sghb_min}] {xgb_results.dat};
                \addplot[
                    color=orange,
                    thick, 
                    smooth
                    ]
                    table [x={n}, y={rs_min}] {xgb_results.dat};
                \addplot[
                    color=purple,
                    thick, 
                    smooth
                    ]
                    table [x={n}, y={bo_min}] {xgb_results.dat};
                \addplot[
                    color=yellow,
                    thick, 
                    smooth
                    ]
                    table [x={n}, y={bohb_min}] {xgb_results.dat};
                \addplot[
                    color=red,
                    thick, 
                    smooth
                    ]
                    table [x={n}, y={smac_min}] {xgb_results.dat};
                    
                    \legend{RGHL,RGHB,BO,BOHB,SMAC}   
            \end{axis}
        \end{tikzpicture}
    }
    \hfill
    \subfloat[Neuralnet Benchmark]{
        \begin{tikzpicture}
            \begin{axis}[
                xlabel={},
                ylabel={},
                legend pos=north east,
                ymajorgrids=true,
                grid style=dashed,
                legend columns=2,
                legend style={font=\tiny},
                tick label style={font=\tiny},
                xlabel style={font=\tiny},
                width=0.28\textwidth,
                height=0.26\textwidth,
            ]
            
                \addplot[
                    color=blue,
                    opacity=1,
                    thick, 
                    smooth
                    ]
                    table [x={n}, y={sghl_min}] {nn_results.dat};
                \addplot[
                    color=green,
                    opacity=1,
                    thick, 
                    smooth
                    ]
                    table [x={n}, y={sghb_min}] {nn_results.dat};
                \addplot[
                    color=orange,
                    thick, 
                    smooth
                    ]
                    table [x={n}, y={rs_min}] {nn_results.dat};
                \addplot[
                    color=purple,
                    thick, 
                    smooth
                    ]
                    table [x={n}, y={bo_min}] {nn_results.dat};
                \addplot[
                    color=yellow,
                    thick, 
                    smooth
                    ]
                    table [x={n}, y={bohb_min}] {nn_results.dat};
                \addplot[
                    color=red,
                    thick, 
                    smooth
                    ]
                    table [x={n}, y={smac_min}] {nn_results.dat};
                \legend{}    
            \end{axis}
        \end{tikzpicture}
    }
    \hfill
    \subfloat[Random Forest Benchmark]{
        \begin{tikzpicture}
            \begin{axis}[
                xlabel={},
                ylabel={},
                legend pos=north east,
                ymajorgrids=true,
                grid style=dashed,
                legend columns=2,
                legend style={font=\tiny},
                tick label style={font=\tiny},
                xlabel style={font=\tiny},
                width=0.28\textwidth,
                height=0.26\textwidth,
            ]
            
                \addplot[
                    color=blue,
                    opacity=1,
                    thick, 
                    smooth
                    ]
                    table [x={n}, y={sghl_min}] {rf_results.dat};
                \addplot[
                    color=green,
                    opacity=1,
                    thick, 
                    smooth
                    ]
                    table [x={n}, y={sghb_min}] {rf_results.dat};
                \addplot[
                    color=orange,
                    thick, 
                    smooth
                    ]
                    table [x={n}, y={rs_min}] {rf_results.dat};
                \addplot[
                    color=purple,
                    thick, 
                    smooth
                    ]
                    table [x={n}, y={bo_min}] {rf_results.dat};
                \addplot[
                    color=yellow,
                    thick, 
                    smooth
                    ]
                    table [x={n}, y={bohb_min}] {rf_results.dat};
                \addplot[
                    color=red,
                    thick, 
                    smooth
                    ]
                    table [x={n}, y={smac_min}] {rf_results.dat};
                \legend{}    
            \end{axis}
        \end{tikzpicture}
    }
    \hfill
    \subfloat[SVM Benchmark]{
        \begin{tikzpicture}
            \begin{axis}[
                xlabel={},
                ylabel={},
                legend pos=north east,
                ymajorgrids=true,
                grid style=dashed,
                legend columns=2,
                legend style={font=\tiny},
                tick label style={font=\tiny},
                xlabel style={font=\tiny},
                width=0.28\textwidth,
                height=0.26\textwidth,
            ]
            
                \addplot[
                    color=blue,
                    thick, 
                    smooth
                    ]
                    table [x={n}, y={sghl_min}] {svm_results.dat};
                \addplot[
                    color=green,
                    thick, 
                    smooth
                    ]
                    table [x={n}, y={sghb_min}] {svm_results.dat};
                \addplot[
                    color=orange,
                    thick, 
                    smooth
                    ]
                    table [x={n}, y={rs_min}] {svm_results.dat};
                \addplot[
                    color=purple,
                    thick, 
                    smooth
                    ]
                    table [x={n}, y={bo_min}] {svm_results.dat};
                \addplot[
                    color=yellow,
                    thick, 
                    smooth
                    ]
                    table [x={n}, y={bohb_min}] {svm_results.dat};
                \addplot[
                    color=red,
                    thick, 
                    smooth
                    ]
                    table [x={n}, y={smac_min}] {svm_results.dat};
                \legend{}    
            \end{axis}
        \end{tikzpicture}
    }
    \caption{Valid loss evaluation per black-box calls(Step-wise average of min.)}
\end{figure*}
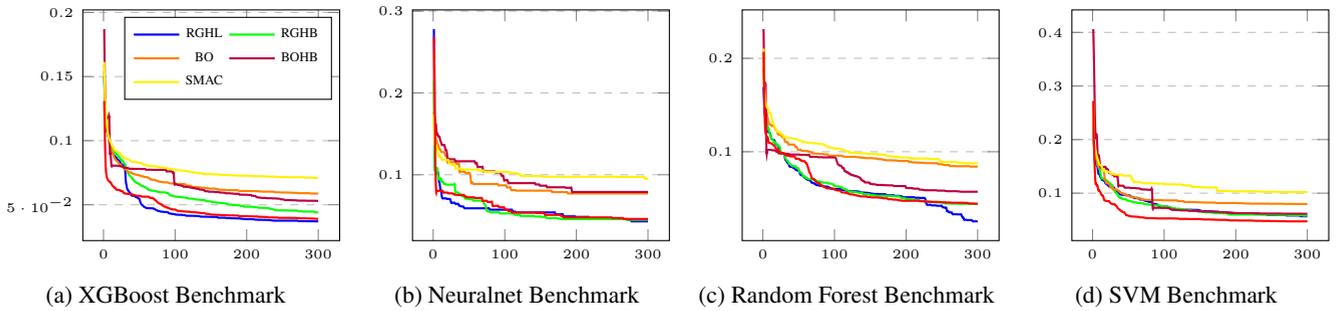

%
%
\begin{figure*}
    \centering
    \subfloat[XGBoost Benchmark]{
        \begin{tikzpicture}
            \begin{axis}[
                ymin=0.07, ymax=0.25,
                legend pos=north east,
                ymajorgrids=true,
                grid style=dashed,
                legend columns=2,
                legend style={font=\tiny},
                tick label style={font=\tiny},
                xlabel style={font=\tiny},
                width=0.28\textwidth,
                height=0.26\textwidth,
            ]
                \addplot[
                    color=blue,
                    opacity=1,
                    thick, 
                    smooth
                    ]
                    table [x={n}, y={sghl_mean}] {xgb_results.dat};
                \addplot[
                    color=green,
                    opacity=1,
                    thick, 
                    smooth
                    ]
                    table [x={n}, y={sghb_mean}] {xgb_results.dat};
                \addplot[
                    color=purple,
                    thick, 
                    smooth
                    ]
                    table [x={n}, y={bo_mean}] {xgb_results.dat};
                \addplot[
                    color=yellow,
                    thick, 
                    smooth
                    ]
                    table [x={n}, y={bohb_mean}] {xgb_results.dat};
                \addplot[
                    color=red,
                    thick, 
                    smooth
                    ]
                    table [x={n}, y={smac_mean}] {xgb_results.dat};
                    
                    \legend{RGHL,RGHB,BO,BOHB,SMAC}
            \end{axis}
        \end{tikzpicture}
    }
    \hfill
    \subfloat[Neuralnet Benchmark]{
        \begin{tikzpicture}

            \begin{axis}[
                xlabel={},
                ylabel={},
                legend pos=north east,
                ymajorgrids=true,
                grid style=dashed,
                legend columns=2,
                legend style={font=\tiny},
                tick label style={font=\tiny},
                xlabel style={font=\tiny},
                width=0.28\textwidth,
                height=0.26\textwidth,
            ]
            
                \addplot[
                    color=blue,
                    opacity=1,
                    thick, 
                    smooth
                    ]
                    table [x={n}, y={sghl_mean}] {nn_results.dat};
                \addplot[
                    color=green,
                    opacity=1,
                    thick, 
                    smooth
                    ]
                    table [x={n}, y={sghb_mean}] {nn_results.dat};
                \addplot[
                    color=purple,
                    thick, 
                    smooth
                    ]
                    table [x={n}, y={bo_mean}] {nn_results.dat};
                \addplot[
                    color=yellow,
                    thick, 
                    smooth
                    ]
                    table [x={n}, y={bohb_mean}] {nn_results.dat};
                \addplot[
                    color=red,
                    thick, 
                    smooth
                    ]
                    table [x={n}, y={smac_mean}] {nn_results.dat};
                \legend{}    
            \end{axis}
        \end{tikzpicture}
    }
    \hfill
    \subfloat[Random Forest Benchmark]{
        \begin{tikzpicture}
            \begin{axis}[
                xlabel={},
                ylabel={},
                legend pos=north east,
                ymajorgrids=true,
                grid style=dashed,
                legend columns=2,
                legend style={font=\tiny},
                tick label style={font=\tiny},
                xlabel style={font=\tiny},
                width=0.28\textwidth,
                height=0.26\textwidth,
            ]
            
                \addplot[
                    color=blue,
                    opacity=1,
                    thick, 
                    smooth
                    ]
                    table [x={n}, y={sghl_mean}] {rf_results.dat};
                \addplot[
                    color=green,
                    opacity=1,
                    thick, 
                    smooth
                    ]
                    table [x={n}, y={sghb_mean}] {rf_results.dat};
                \addplot[
                    color=purple,
                    thick, 
                    smooth
                    ]
                    table [x={n}, y={bo_mean}] {rf_results.dat};
                \addplot[
                    color=yellow,
                    thick, 
                    smooth
                    ]
                    table [x={n}, y={bohb_mean}] {rf_results.dat};
                \addplot[
                    color=red,
                    thick, 
                    smooth
                    ]
                    table [x={n}, y={smac_mean}] {rf_results.dat};
                \legend{}    
            \end{axis}
        \end{tikzpicture}
    }
    \hfill
    \subfloat[SVM Benchmark]{
        \begin{tikzpicture}
            \begin{axis}[
                xlabel={},
                ylabel={},
                legend pos=north east,
                ymajorgrids=true,
                grid style=dashed,
                legend columns=2,
                legend style={font=\tiny},
                tick label style={font=\tiny},
                xlabel style={font=\tiny},
                width=0.28\textwidth,
                height=0.26\textwidth,
            ]
            
                \addplot[
                    color=blue,
                    opacity=1,
                    thick, 
                    smooth
                    ]
                    table [x={n}, y={sghl_mean}] {svm_results.dat};
                \addplot[
                    color=green,
                    opacity=1,
                    thick, 
                    smooth
                    ]
                    table [x={n}, y={sghb_mean}] {svm_results.dat};
                \addplot[
                    color=purple,
                    thick, 
                    smooth
                    ]
                    table [x={n}, y={bo_mean}] {svm_results.dat};
                \addplot[
                    color=yellow,
                    thick, 
                    smooth
                    ]
                    table [x={n}, y={bohb_mean}] {svm_results.dat};
                \addplot[
                    color=red,
                    thick, 
                    smooth
                    ]
                    table [x={n}, y={smac_mean}] {svm_results.dat};
                \legend{}    
            \end{axis}
        \end{tikzpicture}
    }
    \caption{Valid loss evaluation by number of black-box calls(Mean of step-wise cumulative averages)}
\end{figure*}
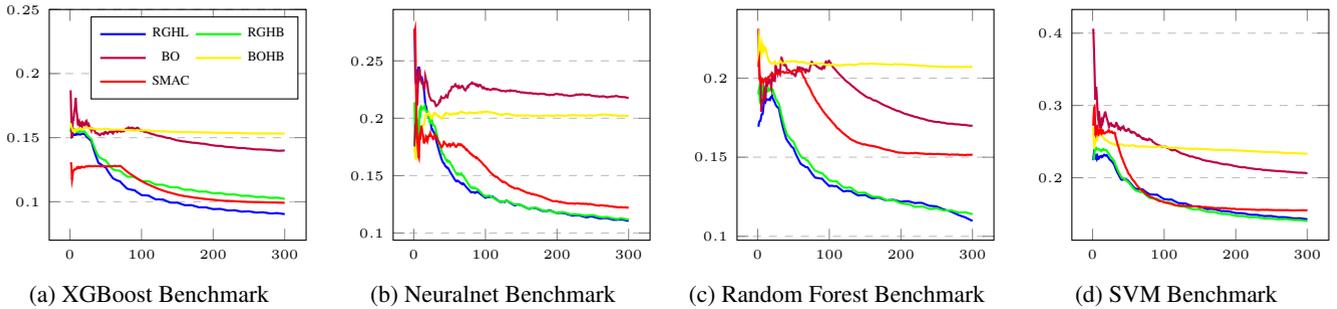

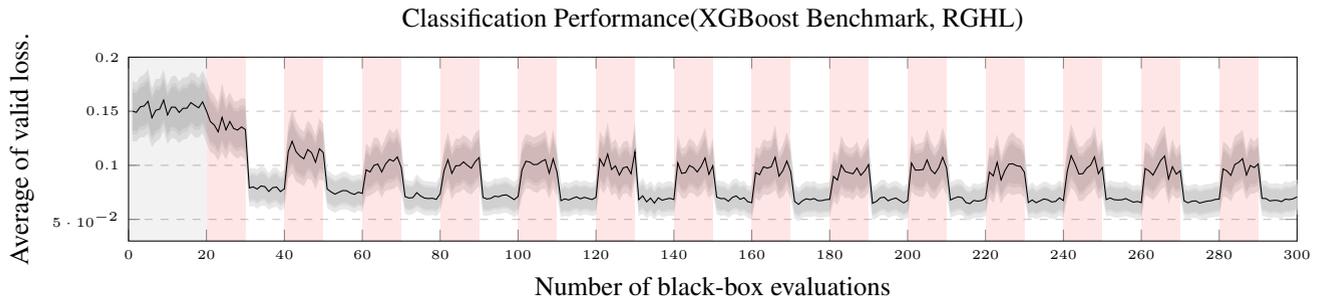
\begin{figure*}[!t]
\centering

\begin{tikzpicture}
    \pgfplotsset{width=17cm, height=4cm,compat=1.8}
    \begin{axis}[domain = 0:300, enlarge x limits = false,
    title={Classification Performance(XGBoost Benchmark, RGHL)},
                xlabel={Number of black-box evaluations},
                ylabel={Average of valid loss.},
                xmin=0, xmax=300,
                ymin=0.03, ymax=0.2,
                legend pos=north east,
                ymajorgrids=true,
                grid style=dashed,
                legend columns=2,
                legend style={font=\tiny},
                tick label style={font=\tiny}
    ]
    
        \begin{scope}
            \fill[gray,opacity=0.1] ({rel axis cs:0,0}) rectangle ({rel axis cs:0.0666,1});
            \fill[red,opacity=0.1] ({rel axis cs:	0.066666667	,0}) rectangle ({rel axis cs:	0.1	,1});
            \fill[white,opacity=0.1] ({rel axis cs:	0.1	,0}) rectangle ({rel axis cs:	0.133333333	,1});
            \fill[red,opacity=0.1] ({rel axis cs:	0.133333333	,0}) rectangle ({rel axis cs:	0.166666667	,1});
            \fill[white,opacity=0.1] ({rel axis cs:	0.166666667	,0}) rectangle ({rel axis cs:	0.2	,1});
            \fill[red,opacity=0.1] ({rel axis cs:	0.2	,0}) rectangle ({rel axis cs:	0.233333333	,1});
            \fill[white,opacity=0.1] ({rel axis cs:	0.233333333	,0}) rectangle ({rel axis cs:	0.266666667	,1});
            \fill[red,opacity=0.1] ({rel axis cs:	0.266666667	,0}) rectangle ({rel axis cs:	0.3	,1});
            \fill[white,opacity=0.1] ({rel axis cs:	0.3	,0}) rectangle ({rel axis cs:	0.333333333	,1});
            \fill[red,opacity=0.1] ({rel axis cs:	0.333333333	,0}) rectangle ({rel axis cs:	0.366666667	,1});
            \fill[white,opacity=0.1] ({rel axis cs:	0.366666667	,0}) rectangle ({rel axis cs:	0.4	,1});
            \fill[red,opacity=0.1] ({rel axis cs:	0.4	,0}) rectangle ({rel axis cs:	0.433333333	,1});
            \fill[white,opacity=0.1] ({rel axis cs:	0.433333333	,0}) rectangle ({rel axis cs:	0.466666667	,1});
            \fill[red,opacity=0.1] ({rel axis cs:	0.466666667	,0}) rectangle ({rel axis cs:	0.5	,1});
            \fill[white,opacity=0.1] ({rel axis cs:	0.5	,0}) rectangle ({rel axis cs:	0.533333333	,1});
            \fill[red,opacity=0.1] ({rel axis cs:	0.533333333	,0}) rectangle ({rel axis cs:	0.566666667	,1});
            \fill[white,opacity=0.1] ({rel axis cs:	0.566666667	,0}) rectangle ({rel axis cs:	0.6	,1});
            \fill[red,opacity=0.1] ({rel axis cs:	0.6	,0}) rectangle ({rel axis cs:	0.633333333	,1});
            \fill[white,opacity=0.1] ({rel axis cs:	0.633333333	,0}) rectangle ({rel axis cs:	0.666666667	,1});
            \fill[red,opacity=0.1] ({rel axis cs:	0.666666667	,0}) rectangle ({rel axis cs:	0.7	,1});
            \fill[white,opacity=0.1] ({rel axis cs:	0.7	,0}) rectangle ({rel axis cs:	0.733333333	,1});
            \fill[red,opacity=0.1] ({rel axis cs:	0.733333333	,0}) rectangle ({rel axis cs:	0.766666667	,1});
            \fill[white,opacity=0.1] ({rel axis cs:	0.766666667	,0}) rectangle ({rel axis cs:	0.8	,1});
            \fill[red,opacity=0.1] ({rel axis cs:	0.8	,0}) rectangle ({rel axis cs:	0.833333333	,1});
            \fill[white,opacity=0.1] ({rel axis cs:	0.833333333	,0}) rectangle ({rel axis cs:	0.866666667	,1});
            \fill[red,opacity=0.1] ({rel axis cs:	0.866666667	,0}) rectangle ({rel axis cs:	0.9	,1});
            \fill[white,opacity=0.1] ({rel axis cs:	0.9	,0}) rectangle ({rel axis cs:	0.933333333	,1});
            \fill[red,opacity=0.1] ({rel axis cs:	0.933333333	,0}) rectangle ({rel axis cs:	0.966666667	,1});
            \fill[white,opacity=0.1] ({rel axis cs:	0.966666667	,0}) rectangle ({rel axis cs:	1	,1});
        
        \end{scope}
            
        \addplot [name path=lower99, fill=none, draw=none] table [
            x=X, y expr=\thisrow{Y} - \thisrow{CI99}]{ci.txt};
        \addplot [name path=upper99, fill=none, draw=none] table [
            x=X, y expr=\thisrow{Y} + \thisrow{CI99}]{ci.txt};
        \addplot[gray!20] fill between[of=lower99 and upper99];
        
        \addplot [name path=lower95, fill=none, draw=none] table [
            x=X, y expr=\thisrow{Y} - \thisrow{CI95}]{ci.txt};
        \addplot [name path=upper95, fill=none, draw=none] table [
            x=X, y expr=\thisrow{Y} + \thisrow{CI95}]{ci.txt};
        \addplot[gray!30] fill between[of=lower95 and upper95];
        
        \addplot [name path=lower90, fill=none, draw=none] table [
            x=X, y expr=\thisrow{Y} - \thisrow{CI90}]{ci.txt};
        \addplot [name path=upper90, fill=none, draw=none] table [
            x=X, y expr=\thisrow{Y} + \thisrow{CI90}]{ci.txt};
        \addplot[gray!40] fill between[of=lower90 and upper90];
        \addplot [] table [x=X, y=Y]{ci.txt};
    
    \end{axis}

\end{tikzpicture}
\caption{The figure visualizes the mean value of the fitness score computed for each step in the XGB benchmark (black line) and its $95\%$ and $99\%$ confidence intervals (gray, light gray area). The first 20 steps in the figure (ultra-light gray background) represent the random population, the red background represents the region of the Rapid Genetic Exploration strategy, and the white background represents the interval of Linear Surrogate Exploitation. The Rapid Genetic Exploration region mainly focused on the uninformative space of the exploration, resulting in lower fitness scores on average. On the other hand, in the Linear Surrogate Exploitation section, we can see that the strategy of targeting high-performing scores led to higher scores.}
\end{figure*}

Figures 6 and 7 show the results of HPOBench experiments based on the OpenML-CC18 dataset. In this experiment, the optimization process is repeated 100 times with 300 black-box function observations for 6 selected HPOs on 4 benchmark environments: XGB, MLP, Random Forest, and SVM.

Figure 6 visualizes the average of the minimum fitness score up to each step, which allows one to easily compare the performance of each HPO. In our experiments, RGHL dominates, but in the SVM benchmark, SMAC outperforms from the beginning to step 300.

Figure 7 visualizes the cumulative average of the fitness score means calculated for each step. This allows for a step-by-step analysis of the difference in overall optimization performance between the HPO strategies. In particular, it can be seen from subgraphs (a), (b), and (c) in Figure 3 that RGHL and SGHB are stable and perform well throughout the entire process. Although SMAC effectively finds the optimal value in the SVM environment, when considering the overall optimization performance as determined by the cumulative average, it is noteworthy that RGHL and SGHB performed better.

%
%
\begin{table}
\caption{Performance of Neuralnet HPOBench}
\label{xgb-perf-table}
\setlength{\tabcolsep}{3pt}
\begin{tabular}{|c|r|r|r|}
\hline
Algorithm& 
Test Loss& 
Valid Loss& 
Time(sec)\\
\hline
RGHL&	0.1774±0.008625&	{\textbf{0.0430±0.003274}}&	12303.170±1477.50701\\
RGHB&	0.1676±0.008550&	0.0449±0.003340&	17752.706±2206.44557\\
RS&	0.3067±0.017417&	0.2876±0.008102&	808.547±0038.85767\\
BO&	0.1537±0.007543&	0.0789±0.005616&	712.431±0025.77543\\
BOHB&	0.1667±0.006390&	0.0950±0.005909&	927.046±0015.13001\\
SMAC&	0.1677±0.008242&	0.0459±0.003699&	5577.922±0627.60109\\

\hline
\end{tabular}
\end{table}

%
%
\begin{table}
\caption{Performance of XGB HPOBench}
\label{xgb-perf-table}
\setlength{\tabcolsep}{3pt}
\begin{tabular}{|c|r|r|r|}
\hline
Algorithm& 
Test Loss& 
Valid Loss& 
Time(sec)\\
\hline
RGHL&	0.1858±0.007484&	{\textbf{0.0372±0.002226}}&	419.962484±44.952676\\
RGHB&	0.1787±0.007153&	0.0441±0.002588&	442.980967±45.393322\\
RS&	0.1851±0.007007&	0.6696±0.009390&	206.583917±20.912707\\
BO&	0.1829±0.007253&	0.0530±0.002928&	326.795026±27.579218\\
BOHB&	0.1776±0.007311&	0.0708±0.003317&	793.597957±21.065918\\
SMAC&	0.1820±0.007156&	0.0390±0.002341&	416.835349±34.106769\\

\hline
\end{tabular}
\end{table}

%
%
\begin{table}
\caption{Performance of RF HPOBench}
\label{xgb-perf-table}
\setlength{\tabcolsep}{3pt}
\begin{tabular}{|c|r|r|r|}
\hline
Algorithm& 
Test Loss& 
Valid Loss& 
Time(sec)\\
\hline
RGHL&	0.1684±0.006773 &	0.0412±0.002238 &	123.597146±6.188801 \\
RGHB&	0.1667±0.006502 &	{\textbf{0.0407±0.002348 }}&	147.708267±5.552665 \\
RS&	0.2102±0.008689 &	0.5008±0.016402 &	121.677629±2.104253 \\
BO&	0.1712±0.006910 &	0.0527±0.002758 &	221.799096±5.050753 \\
BOHB&	0.1690±0.006763 &	0.0818±0.003877 &	672.364848±2.492630 \\
SMAC&	0.1708±0.006713 &	0.0411±0.002348 &	334.240021±6.427550 \\

\hline
\end{tabular}
\end{table}
%
%
\begin{table}
\caption{Performance of SVM HPOBench}
\label{xgb-perf-table}
\setlength{\tabcolsep}{3pt}
\begin{tabular}{|c|r|r|r|}
\hline
Algorithm& 
Test Loss& 
Valid Loss& 
Time(sec)\\
\hline
RGHL&	0.3009±0.013811 &	0.0548±0.003662 &	14.949991±1.226181 \\
RGHB&	0.3084±0.015773 &	0.0708±0.004270 &	74.199489±3.253045 \\
RS&	0.4534±0.014416 &	0.5351±0.013855 &	14.991651±0.953568 \\
BO&	0.2077±0.007916 &	0.0569±0.002816 &	62.234121±1.384060 \\
BOHB&	0.2378±0.010158 &	0.0933±0.003862 &	563.529929±1.172031 \\
SMAC&	0.1821±0.008033 &	{\textbf{0.0529±0.002517 }}&	118.696979±1.012420\\

\hline
\end{tabular}
\end{table}

Tables 3-6 present the results of evaluating the two techniques proposed in this work and the four existing HPO methods on four benchmark environments in HPOBench. For each benchmark environment, the experiments were conducted using the dataset in Table 2, calling the black-box function 300 times and training the model with the optimal hyperparameters obtained, and the average effective loss and test loss were analyzed. The results are presented in detail, including the mean values and 95\% confidence intervals around the test loss, effective loss, and time taken for each technique. The results showed that the RGHL strategy performed well in most cases, but was observed to be significantly more time-consuming on Neuralnet HPOBench. 

\section{Conclusion}
This paper presents a novel approach to solving the HPO problem. One of the main problems with existing Bayesian-based HPO is that the trade-off between exploration and exploitation must be set empirically. To overcome this limitation, we propose a method that ensures equal competition between exploration and exploitation.
Specifically, this approach combines an exploration strategy using a radical GA with an exploitation strategy using a robust linear surrogate model to ensure that the two strategies compete with equal weight. This approach effectively prevents the optimization process from converging to local optima. Therefore, this method not only improves the performance in the process of finding the optimal value but also plays an important role in increasing the efficiency and effectiveness of HPO strategies.

The overall experimental results show that in the early stages of black-box function calls, the Bayesian HPO strategies outperformed the RGHL introduced in this paper. However, as the experimental steps progressed, we observed that these strategies tended to get stuck in local optima, and most of the RGHLs outperformed on average over 50 steps, which is one of the important findings of this study.

In the NeuralNet HPOBench, we identified that the time delay problem in RGHL arises because the RHC, employed as the acquisition function for surrogate model exploitation, fails to converge to the optimal value and repeatedly initializes itself in random directions, resulting in significant delays. We identified the need for a fast termination condition to mitigate the problem. However, this issue requires a more fundamental solution beyond incremental improvements. We plan to explore this challenge in greater depth as the primary focus of our next study.

For future research, we plan to further explore the time delay and the “Time-division strategy”, which dynamically applies different HPO strategies per step. This will further improve the RGHL HPO strategy and maximize its performance and reliability. We anticipate that our research will make significant contributions to advancing the efficiency and effectiveness of HPO methodologies.

\vspace{-300pt}
\begin{IEEEbiography}[{\includegraphics[width=1in,height=1.25in,clip,keepaspectratio]{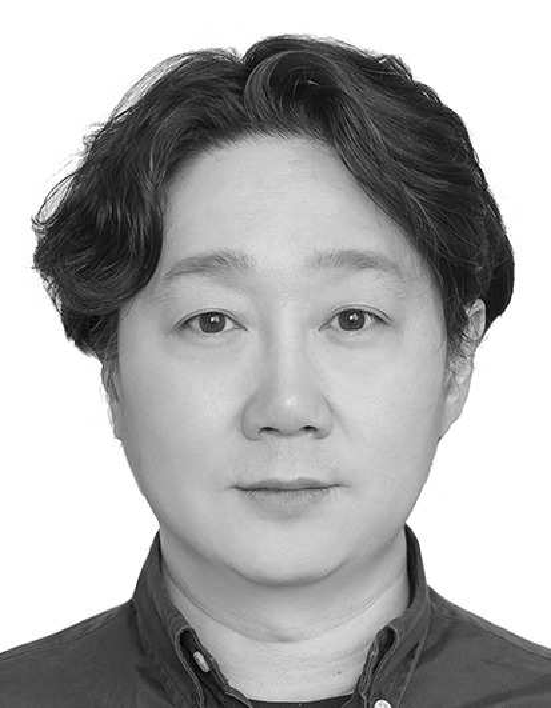}}]{Chul Kim} received the M.S. degree in Graduate School of Engineering of Computer Science \& Engineering from Hanyang University, Seoul, Korea, in 2013, respectively.
He is currently a Director with the AI Research Center, Thingspire, Seoul, Korea, and was an Adjunct Professor with the Graduate School of Engineering of Computer Science \& Engineering, Hanyang University, Seoul, Korea, in 2018, 2020 and 2022.
His research interests include forecasting, deep reinforcement learning, optimization, computer vision, and anomaly detection in industrial domains.
\end{IEEEbiography}
\noindent

\vspace{-300pt} 

\begin{IEEEbiography}[{\includegraphics[width=1in,height=1.25in,clip,keepaspectratio]{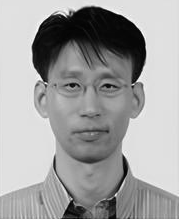}}]{Inwhee Joe} received his BS and MS in electronics engineering from Hanyang University, Seoul, Korea, and his Ph.D. in electrical and computer engineering from Georgia Institute of Technology, Atlanta, GA in 1998. Since 2002, he has been a faculty member in the Department of Computer Science at Hanyang University, Seoul, Korea. His current research interests include the Internet of Things, cellular systems, wireless-powered communication networks, embedded systems, network security, machine learning, and performance evaluation.
\end{IEEEbiography}

\EOD

\end{document}